\def\year{2020}\relax
\documentclass[letterpaper]{article} 
\usepackage{aaai20}  
\usepackage{times}  
\usepackage{helvet} 
\usepackage{courier}  
\usepackage[hyphens]{url}  
\usepackage{graphicx} 
\usepackage{booktabs}
\usepackage{amssymb}
\usepackage{pifont}
\usepackage{xcolor}
\usepackage{multirow,subfig}

\usepackage{changepage}
\usepackage{amsthm}
\usepackage{array}
\usepackage{amsmath}
\usepackage{bm}
\usepackage{enumitem}
\theoremstyle{definition}
\newtheorem{definition}{Definition}

\newcommand{\cmark}{\ding{51}}%
\newcommand{\xmark}{\ding{55}}%
\usepackage{graphicx}  
\newcommand{\proposed}{\textsf{DMGI}}
\newcommand{\proposedattn}{\textsf{DMGI}$_{\textsf{attn}}$}

\newcommand{\citet}[1]{\citeauthor{#1} \shortcite{#1}}
\nocopyright
\title{Unsupervised Attributed Multiplex Network Embedding}
\author{Chanyoung Park\textsuperscript{\rm 1}, Donghyun Kim\textsuperscript{\rm 2}, Jiawei Han\textsuperscript{\rm 1}, Hwanjo Yu\textsuperscript{\rm 3} \\
\textsuperscript{\rm 1}Department of Computer Science, University of Illinois at Urbana-Champaign, IL, USA\\
\textsuperscript{\rm 2}Yahoo! Research, CA, USA\\
\textsuperscript{\rm 3}Department of Computer Science and Engineering, Pohang University of Science and Technology, Korea \\
pcy1302@illinois.edu, donghyun.kim@verizonmedia.com, hanj@illinois.edu, hwanjoyu@postech.ac.kr
}

\begin{document}

\maketitle

\begin{abstract}
Nodes in a multiplex network are connected by multiple types of relations. However, most existing network embedding methods assume that only a single type of relation exists between nodes. Even for those that consider the multiplexity of a network, they overlook node attributes, resort to node labels for training, and fail to model the global properties of a graph. We present a simple yet effective unsupervised network embedding method for attributed multiplex network called~\proposed, inspired by Deep Graph Infomax (DGI) that maximizes the mutual information between local patches of a graph, and the global representation of the entire graph. We devise a systematic way to jointly integrate the node embeddings from multiple graphs by introducing 1) the consensus regularization framework that minimizes the disagreements among the relation-type specific node embeddings, and 2) the universal discriminator that discriminates true samples regardless of the relation types.
We also show that the attention mechanism infers the importance of each relation type, and thus can be useful for filtering unnecessary relation types as a preprocessing step. Extensive experiments on various downstream tasks demonstrate that~\proposed~outperforms the state-of-the-art methods, even though~\proposed~is fully unsupervised.
\end{abstract}

\section{Introduction}
Analyzing and mining useful knowledge in graphs have been an actively researched topic for decades both in academia and industry. Among various graph mining techniques, network embedding, which learns low-dimensional vector representations for nodes in a graph, is shown to be especially effective for various network-based tasks~\cite{tang2015line,wang2017community,meng2019co}.


However, most existing network embedding methods assume that only a single type of relation exists between nodes~\cite{velivckovic2017graph,velivckovic2018deep,kipf2016semi}, whereas in reality networks are \textit{multiplex}~\cite{de2013mathematical} in nature, i.e., with multiple types of relations. Taking the publication network as an example, two papers can be connected due to various reasons, such as authors (two papers are authored by a common author), citation (one paper cites the other), or keywords (two papers share common keywords).  As another example, in a movie database network, two movies can be connected via a common director, or a common actor. 

Although different types of relations can independently form different graphs, \textit{these graphs are related}, and thus can mutually help each other for various downstream tasks. As a concrete example of the publication network, although it is hard to infer the topic of a paper only from its citations (citations can be diverse), also knowing other papers written by the same authors will help predict its topic, because authors usually work on a specific research topic.
Furthermore, nodes in graphs may contain attribute information, which plays important roles in many applications~\cite{zhang2018anrl}. For example, if we are additionally given the abstract of the papers in the publication network, it will be much easier to infer their topics. As such, the main challenge is to learn a consensus representation of a node that not only considers its \textit{{multiplexity}}, but also its \textit{{attributes}}.

Several recent studies have been conducted for multiplex network embedding, however, some issues remain that need further consideration. First, previous methods~\cite{qu2017attention,zhang2018scalable,shi2018mvn2vec,liu2017principled} focus on the integration of multiple graphs, but overlook \textbf{node attributes}. Second, even for those that consider node attributes~\cite{schlichtkrull2018modeling,wang2019heterogeneous}, they require \textbf{node labels} for training. However, as node labeling is often expensive and time-consuming, it would be the best if a method can show competitive performance even without any label. 
Third, most of these methods fail to model the \textbf{global properties} of a graph, because they are based on random walk-based skip-gram model or graph convolutional network (GCN)~\cite{kipf2016semi}, both of which are known to be effective for capturing the local graph structure~\cite{yadav2019lovasz}. More precisely, nodes that are ``close'' (i.e., within the same context window or neighborhoods) in the graph are trained to have similar representations, whereas nodes that are far apart do not have similar representations, even though they are structurally similar~\cite{ribeiro2017struc2vec}.

Keeping these limitations in mind, we propose a simple yet effective unsupervised method for embedding attributed multiplex networks. The core building block of our proposed method is Deep Graph Infomax (DGI)~\cite{velivckovic2018deep} that aims to learn a node encoder that maximizes the mutual information between local patches of a graph, and the global representation of the entire graph.
DGI is the workhorse method for our task, because it 1) naturally integrates the node attributes by using a GCN, 2) is trained in a fully unsupervised manner, and 3) captures the global properties of the entire graph. However, it is challenging to apply DGI, which is designed for embedding a single network, to a multiplex network in which the interactions among multiple relation types, and the importance of each relation type should be considered.

In this paper, we present a systematic way to jointly integrate the embeddings from multiple types of relations between nodes, so as to facilitate them to mutually help each other learn high-quality embeddings useful for various downstream tasks. More precisely, we introduce the \textit{consensus regularization framework} that minimizes the disagreements among the relation-type specific node embeddings, and the \textit{universal discriminator} that discriminates true samples, i.e., ground truth ``(graph-level summary, local patch)'' pairs, regardless of the relation types. Moreover, we demonstrate that through the \textit{attention mechanism}, we can infer the importance of each relation type in generating the consensus node embeddings, which can be used for filtering unnecessary relation types as a preprocessing step.
Our extensive experiments demonstrate that our proposed method, Deep Multplex Graph Infomax (\proposed), outperforms the state-of-the-art attributed multiplex network embedding methods in terms of node clustering, similarity search, and especially, node classification even though~\proposed~is fully unsupervised. 

\section{Problem Statement}
\begin{definition}
	\textbf{(Attributed Multiplex Network)} An attributed multiplex network is a network $\mathcal{G}=\{\mathcal{G}^1,\mathcal{G}^2,...,\mathcal{G}^{|\mathcal{R}|}\}=\{\mathcal{V},\mathcal{E},\textbf{X}\}$, where $\mathcal{G}^r=\{\mathcal{V},\mathcal{E}^{(r)},\textbf{X}\}$ is a graph of the relation type $r\in\mathcal{R}$,
	$\mathcal{V}$ is the set of $n$ nodes, $\mathcal{E}=\bigcup_{r\in \mathcal{R}} \mathcal{E}^{(r)}\subseteq \mathcal{V}\times\mathcal{V}$ is the set of all edges with relation type $r\in\mathcal{R}$, and ${\textbf{X}}\in\mathbb{R}^{n \times f}$ is a matrix that encodes node attributes information for $n$ nodes. Note that $|\mathcal{R}|>1$ for multiplex networks, and $|\mathcal{R}|=1$ for a single network. Given the network $\mathcal{G}$, $\mathcal{A}=\{\textbf{A}^{(1)},...,\textbf{A}^{(|\mathcal{R}|)}\}$ is a set of adjacency matrices, where $\textbf{A}^{(r)}\in\{0,1\}^{|V|\times|V|}$ is an adjacency matrix of the network $\mathcal{G}^r$.
\end{definition}

\smallskip
\noindent\textbf{Task:}
	\textbf{Unsupervised Attributed Multiplex Network Embedding. } Given an attributed multiplex network $\mathcal{G}=\{\mathcal{V},\mathcal{E},\textbf{X}\}$, and the set of adjacency matrices $\mathcal{A}$, the task of unsupervised attributed multiplex network embedding is to learn a $d$-dimensional vector representation $\textbf{z}_i\in\mathcal{\textbf{Z}}\in\mathbb{R}^{n\times d}$ for each node $v_i\in \mathcal{V}$ without using any labels.

\section{Unsupervised Attributed Multiplex Network Embedding}
We begin by introducing Deep Graph Informax (DGI)~\cite{velivckovic2018deep}, then we discuss about its limitations, and present our proposed method.

\smallskip
\noindent\textbf{Deep Graph Infomax (DGI).}
\citet{velivckovic2018deep} proposed an unsupervised method for learning node representations, called DGI, that relies on the infomax principle~\cite{linsker1988self}. More precisely, DGI aims to learn a low-dimensional vector representation for each node $v_i$, i.e., $\textbf{h}_i\in\mathbb{R}^d$, such that the average mutual information (MI) between the graph-level (global) summary representation $\textbf{s}\in\mathbb{R}^d$, and the representations of the local patches $\{\textbf{h}_1,\textbf{h}_2,...,\textbf{h}_n\}$ is maximized. To this end, DGI introduces a discriminator $\mathcal{D}$ that discriminates the true samples, i.e., $(\textbf{h}_i, \textbf{s})$, from its negative counterparts, i.e., $(\bm{\tilde{\textbf{h}}}_j,\textbf{s})$:
\begin{equation}
\begin{split}
\mathcal{L}=\sum_{v_i\in\mathcal{V}}^{n} \log \mathcal{D}\left({\textbf{h}}_{i}, {\textbf{s}}\right)+\sum_{j=1}^{n}\log \left(1-\mathcal{D}\left({\bm{\tilde{\textbf{h}}}_j}, {\textbf{s}}\right)\right)
\end{split}
\label{eqn:dgi}
\end{equation}
where $\textbf{h}_i=\sigma\left(\sum_{j\in N(i)} \frac{1}{c_{i j}}  \textbf{x}_j\textbf{W}\right)$, $N(i)$ is the set of neighboring nodes of $v_i$ including $v_i$ itself, $\textbf{W}\in\mathbb{R}^{f\times d}$, and $c_{ij}$ is a normalizing constant for edge $(v_i,v_j)$, $\textbf{s}=\sigma\left(\frac{1}{n} \sum_{i=1}^{n} \textbf{h}_{i}\right)$, and $\sigma$ is the sigmoid nonlinearity.
Negative patch representation $\bm{\tilde{\textbf{h}}}_j$ is obtained by row-wise shuffling the original attribute matrix $\textbf{X}$.
\citet{velivckovic2018deep} theoretically proved that the binary cross entropy loss shown in Eqn.~\ref{eqn:dgi} amounts to maximizing the mutual information (MI) between $\textbf{h}_i$ and $\textbf{s}$, based on the Jensen-Shannon divergence~\cite{velivckovic2018deep}. Refer to Section 3.3 of~\cite{velivckovic2018deep}~for the detailed proof.
As the local patch representations $\{\textbf{h}_1,\textbf{h}_2,...,\textbf{h}_n\}$ are learned to preserve the MI with the graph-level representation $\textbf{s}$, each $\textbf{h}_i$ is expected to capture the global properties of the entire graph.

\smallskip
\noindent\textbf{Limitation.}
Despite its effectiveness, DGI is designed for a single attributed network, and thus it is not straightforward to apply it to a multiplex network. As a naive extension of DGI to a multiplex attributed network, we can independently apply DGI to each graph formed by each relation type, and then compute the average of the embeddings obtained from each graph to get the final node representations.
However, we argue that this fails to model the multiplexity of the network, because the interactions among the node embeddings from different relation types is not captured.
Thus, we need a more systematic way to integrate multiple independent models to obtain the final consensus embedding that every model can agree on.

\subsection{Deep Multiplex Graph Infomax:~\proposed}
We present our unsupervised method for embedding an attributed multiplex network. 
We first describe how to independently model each graph pertaining to each relation type, then explain how to jointly integrate them to finally obtain the consensus node embedding matrix.

\smallskip
\noindent\textbf{Relation-type specific Node Embedding. }
For each relation type $r\in\mathcal{R}$, we introduce a relation-type specific node encoder $g_{r} : \mathbb{R}^{n \times f} \times \mathbb{R}^{n \times n} \rightarrow \mathbb{R}^{n \times d}$ to generate the relation-type specific node embedding matrix $\textbf{H}^{(r)}$ of nodes in $\mathcal{G}^{(r)}$. The encoder is a single--layered GCN:
\begin{equation}
\small
\textbf{H}^{(r)}=g_{r}(\mathbf{X}, \mathbf{A}^{(r)}|\textbf{W}^{(r)})=\sigma\left(\hat{\mathbf{D}}_r^{-\frac{1}{2}} \hat{\mathbf{A}}^{(r)} \hat{\mathbf{D}}_r^{-\frac{1}{2}} \mathbf{X} \textbf{W}^{(r)}\right)
\label{eqn:gcn}
\end{equation}
where $\hat{\mathbf{A}}^{(r)}=\mathbf{A}^{(r)}+w\mathbf{I}_{n}$, $\hat{D}_{i i}=\sum_{j} \hat{A}_{i j}$, $\textbf{W}^{(r)}\in\mathbb{R}^{f\times d}$ is a trainable weight matrix of the relation-type specific decoder $g_r$, and $\sigma$ is the ReLU nonlinearity. Unlike conventional GCNs~\cite{kipf2016semi}, we control the weight of the self-connections by introducing a weight $w\in\mathbb{R}$. Larger $w$ indicates that the node itself plays a more important role in generating its embedding, which in turn diminishes the importance of its neighboring nodes.
Then, we compute the graph-level summary representation $\textbf{s}^{(r)}$ that summarizes the global content of the graph $\mathcal{G}^{(r)}$. We employ a readout function $\textsf{Readout}:\mathbb{R}^{n\times d} \rightarrow \mathbb{R}^d$:
\begin{equation}
\small
\textbf{s}^{(r)}=\textsf{Readout}(\mathbf{H}^{(r)})=\sigma\left(\frac{1}{n} \sum_{i=1}^{n} \textbf{h}^{(r)}_{i}\right)
\label{eqn:readout}
\end{equation}
where $\sigma$ is the logistic sigmoid nonlinearity, and $\textbf{h}_i^{(r)}$ denotes the $i$-th row vector of the matrix $\textbf{H}^{(r)}$. \textcolor{black}{We also note that various pooling methods such as maxpool, and SAGPool~\cite{lee2019self} can be used as {$\textsf{Readout}(\cdot)$}.}

Next, given the relation-type specific node embedding matrix $\textbf{H}^{(r)}$, and its graph-level summary representation $\textbf{s}^{(r)}$, we compute the relation-type specific cross entropy:
\begin{equation}
\small
\begin{split}
\mathcal{L}^{(r)}=\sum_{v_i\in\mathcal{V}}^{n} \log \mathcal{D}\left({\textbf{h}}^{(r)}_{i}, {\textbf{s}^{(r)}}\right)+\sum_{j=1}^{n}\log \left(1-\mathcal{D}\left({\bm{\tilde{\textbf{h}}}}^{(r)}_{j}, {\textbf{s}^{(r)}}\right)\right)
\end{split}
\raisetag{2.7\baselineskip}
\label{eqn:rel_loss}
\end{equation}
where $\mathcal{D}:\mathbb{R}^d \times \mathbb{R}^d \rightarrow \mathbb{R}$ is a discriminator that scores patch-summary representation pairs, i.e., $(\textbf{h}^{(r)}_i,\textbf{s}^{(r)})$. In this paper, we apply a simple bilinear scoring function as it empirically performs the best in our experiments:
\begin{equation}
\small
\mathcal{D}\left({\textbf{h}}^{(r)}_{i}, {\textbf{s}^{(r)}}\right) = \sigma(\textbf{h}_i^{(r)T}\textbf{M}^{(r)}\textbf{s}^{(r)})
\label{eqn:disc}
\end{equation}
where $\sigma$ is the logistic sigmoid nonlinearity, and $\textbf{M}^{(r)}\in\mathbb{R}^{d\times d}$ is a trainable scoring matrix. To generate the negative node embedding $\bm{\tilde{\textbf{h}}}_j^{(r)}$, we corrupt the original attribute matrix by shuffling it in the row-wise manner~\cite{velivckovic2018deep}, i.e., $\bm\tilde{\textbf{X}}\leftarrow \textbf{X}$, and reuse the encoder in Eqn.~\ref{eqn:gcn}. i.e. $\bm\tilde{\textbf{H}}^{(r)} = g_{r}(\bm\tilde{\textbf{X}}, \mathbf{A}^{(r)}|\textbf{W}^{(r)})$.

\subsubsection{Joint Modeling and Consensus Regularization. }
Heretofore, by independently maximizing the average MI between the local patches $\{\textbf{h}^{(r)}_1,\textbf{h}^{(r)}_2,...,\textbf{h}^{(r)}_n\}$ and the graph-level summary $\textbf{s}^{(r)}$ pertaining to each graph $\mathcal{G}^{(r)} (\forall r\in\mathcal{R})$, we obtained relation-type specific node embedding matrix $\textbf{H}^{(r)}$ that captures the global information in $\mathcal{G}^{(r)}$. However, as each $\textbf{H}^{(r)}$ is trained independently for each $r\in\mathcal{R}$, these embedding matrices only contain relevant information regarding each relation type, and therefore fail to take advantage of the multiplexity of the network. This motivates us to develop a systematic way to jointly integrate the embeddings from different relation types, so as to facilitate them to mutually help each other learn high-quality embeddings. 

To this end, we introduce the consensus embedding matrix $\textbf{Z}\in\mathbb{R}^{n\times d}$ on which every relation-type specific node embedding matrix $\textbf{H}^{(r)}$ can agree. More precisely, we introduce the \textit{consensus regularization} framework that consists of 1) a regularizer minimizing the disagreements between the set of original node embeddings, i.e. $\{\textbf{H}^{(r)}\;|\;r\in\mathcal{R} \}$ and the consensus embedding $\textbf{Z}$, and 2) another regularizer maximizing the disagreement between the corrupted node embeddings, i.e., $\{\bm{\tilde{\textbf{H}}}^{(r)}\;|\;r\in\mathcal{R} \}$, and the consensus embedding $\textbf{Z}$, which are formulated as follows:

\begin{equation}
\small
\begin{split}
\ell_{\text{cs}}=
\left[\textbf{Z}-\mathcal{Q}\left(\{\textbf{H}^{(r)}\;|\;r\in\mathcal{R} \}\right)\right]^2-
\left[\textbf{Z}-\mathcal{Q}\left(\{\bm{\tilde{\textbf{H}}}^{(r)}\;|\;r\in\mathcal{R} \}\right)\right]^2
\end{split}
\raisetag{2.5\baselineskip}
\label{eqn:reg}
\end{equation}
where $\mathcal{Q}$ is an aggregation function that combines a set of node embedding matrices from multiple relation types into a single embedding matrix. i.e., $\textbf{H}\in\mathbb{R}^{n\times d}$. 
$\mathcal{Q}$ can be any pooling method that can handle permutation invariant input, such as set2set~\cite{vinyals2015order} or Set Transformer~\cite{lee2018set}. However, considering the efficiency of the method, we simply employ average pooling, i.e., computing the average of the set of embedding matrices:
\begin{equation}
\small
\textbf{H}=\mathcal{Q}\left(\{\textbf{H}^{(r)}\;|\;r\in\mathcal{R}\}\right)=\frac{1}{|\mathcal{R}|}\sum_{r\in\mathcal{R}}\textbf{H}^{(r)}
\label{eqn:avg}
\end{equation}

\noindent It is important to note that the scoring matrix $\textbf{M}^{(*)}$ in Eqn.~\ref{eqn:disc} is shared among all the relations $r\in\mathcal{R}$. i.e., $\textbf{M}=\textbf{M}^{(1)}=\textbf{M}^{(2)}=...=\textbf{M}^{(|\mathcal{R}|)}$. The intuition is to learn the \textit{universal discriminator} that is capable of scoring the true pairs higher than the negative pairs regardless the relation types. We argue that the universal discriminator facilitates the joint modeling of different relation types together with the consensus regularization.

\begin{figure}
	\centering
	\includegraphics[width=\linewidth]{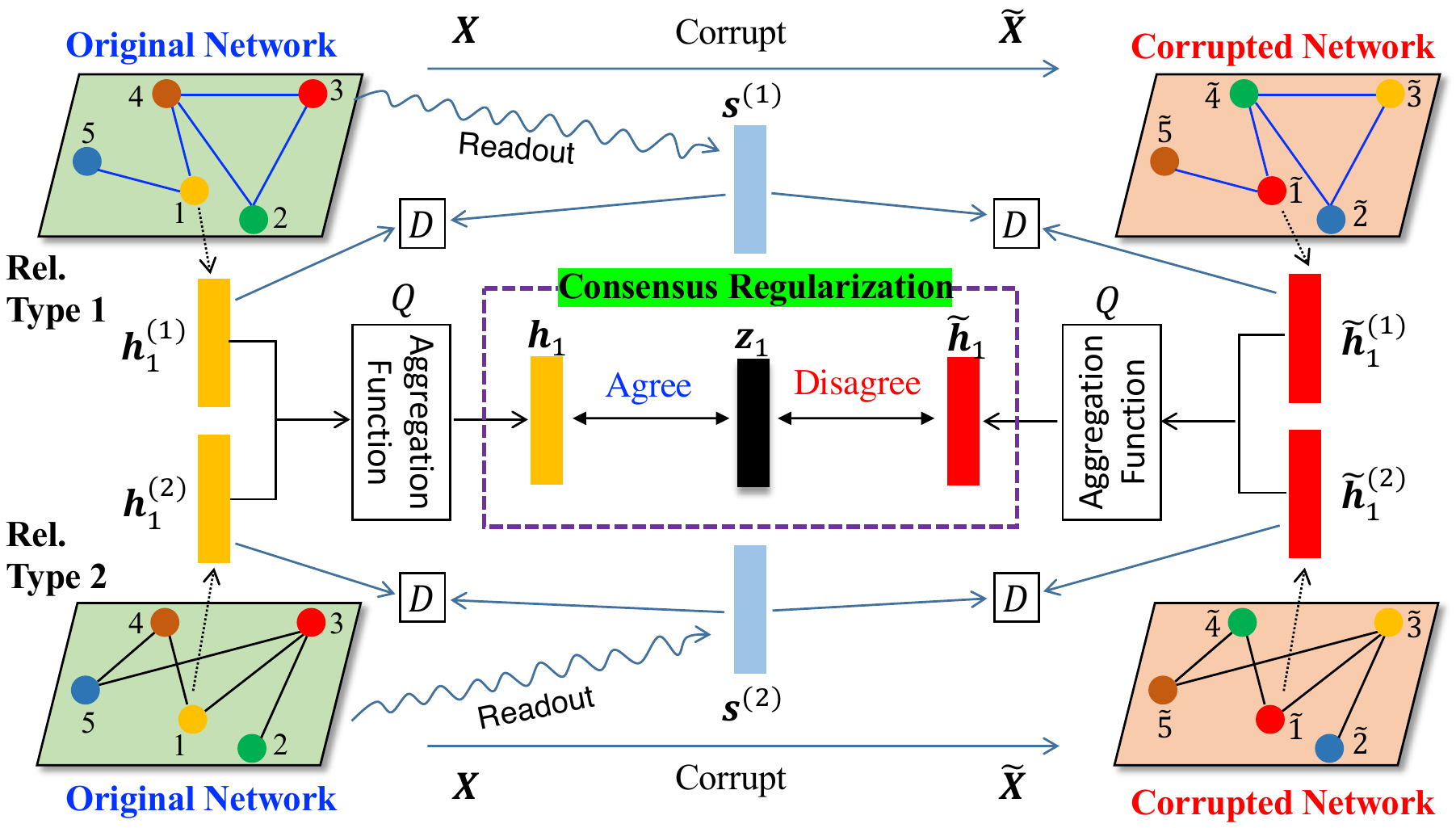}
	\caption{Overview of~\proposed~(Best viewed in color).}
	\label{fig:overall}	
\end{figure}

Finally, we jointly optimize the sum of all the relation-type specific loss in Eqn.~\ref{eqn:rel_loss}, and the consensus regularization in Eqn.~\ref{eqn:reg} to obtain the final objective $\mathcal{J}$ as follows:
\begin{equation}
\mathcal{J}=\sum_{r\in\mathcal{R}}\mathcal{L}^{(r)} + \alpha\ell_{\text{cs}} + \beta||\Theta||^2
\label{eqn:final_loss}
\end{equation}
where $\alpha$ controls the importance of the consensus regularization, $\beta$ is a coefficient for l2 regularization on $\Theta$, which is a set of trainable parameters. i.e., $\Theta=\{\{\textbf{W}^{(r)}\;|\;{r\in\mathcal{R}}\}, \textbf{M}, \textbf{Z}\}$, and $\mathcal{J}$ is optimized by Adam optimizer. Figure~\ref{fig:overall} illustrates the overview of~\proposed.

\begin{table*}[t]
	\centering
	\small
	\caption{Statistics of the datasets. The node attributes are bag-of-words of text associated with each node.}
	\renewcommand{\arraystretch}{0.92}
	\begin{tabular}{c||c|>{\centering\arraybackslash}m{0.7cm}|>{\centering\arraybackslash}m{0.7cm}|>{\centering\arraybackslash}m{0.8cm}||>{\centering\arraybackslash}m{2.2cm}|c|c|c|c}
		& \begin{tabular}[x]{@{}c@{}}Relations \\(A-B)\end{tabular}   & Num. A   & Num. B & Num. A-B & Relation type & \begin{tabular}[x]{@{}c@{}}Num. \\relations\end{tabular} & \begin{tabular}[x]{@{}c@{}}Num. \\node attributes\end{tabular}  &  \begin{tabular}[x]{@{}c@{}}Num. \\labeled data\end{tabular}       & \begin{tabular}[x]{@{}c@{}}Num.  \\classes\end{tabular}   \\
		\hline\hline
		\multirow{2}{*}{ACM}  & \underline{P}aper-\underline{A}uthor   & 3,025  & 5,835    & 9,744  & \underline{P}-\underline{A}-\underline{P} & 29,281  & \multirow{2}{*}{\begin{tabular}[x]{@{}c@{}}1,830\vspace{-0.5ex} \\(Paper abstract)\end{tabular}}  & \multirow{2}{*}{600}                      & \multirow{2}{*}3    \\
		& \underline{P}aper-\underline{S}ubject  & 3,025  & 56       & 3,025  & \underline{P}-\underline{S}-\underline{P}  & 2,210,761 &                                   &                                 &  \\
		\hline
		\multirow{2}{*}{IMDB} & \underline{M}ovie-\underline{A}ctor    & 3,550  & 4,441    & 10,650  & \underline{M}-\underline{A}-\underline{M}  & 66,428  & \multirow{2}{*}{\begin{tabular}[x]{@{}c@{}}1,007\vspace{-0.5ex} \\(Movie plot)\end{tabular}}           & \multirow{2}{*}{300}                & \multirow{2}{*}3 \\
		& \underline{M}ovie-\underline{D}irector & 3,550  & 1,726    & 3,550  & \underline{M}-\underline{D}-\underline{M}   & 13,788  &                                   &                                   & \\
		\hline
		\multirow{3}{*}{DBLP} & \underline{P}aper-\underline{A}uthor   & 7,907  & 1,960    & 14,238 & \underline{P}-\underline{A}-\underline{P}  & 144,783 & \multirow{3}{*}{\begin{tabular}[x]{@{}c@{}}2,000 \\(Paper abstract)\end{tabular}}            & \multirow{3}{*}{80}              & \multirow{3}{*}4   \\
		& \underline{P}aper-\underline{P}aper    & 7,907  & 7,907    & 10,522   & \underline{P}-\underline{P}-\underline{P}   & 90,145  &                                   &                               &  \\
		& \underline{A}uthor-\underline{T}erm    & 1,960  & 1,975    & 57,269  & \underline{P}-\underline{A}-\underline{T}-\underline{A}-\underline{P}   & 57,137,515    &                                &                             & \\
		\hline
		\multirow{3}{*}{Amazon} & \multirow{3}{*}{Item-Item}  & \multirow{3}{*}{7,621} & \multirow{3}{*}{7,621}   & 38,514  & Also-view & 266,237 & \multirow{3}{*}{\begin{tabular}[x]{@{}c@{}}2,000 \\(Item description)\end{tabular}}  & \multirow{3}{*}{80}                          & \multirow{3}{*}4 \\
		&   &  &     &45,446 & Also-bought  & 1,104,257  &                                 &                                 & \\
		&       &  &    & 9,783 & Bought-together  & 16,305   &                                  &                             
	\end{tabular}
	\label{tab:stats}
\end{table*}

\medskip
\noindent\textbf{Discussion. }
Despite its efficiency, the above average pooling scheme in Eqn.~\ref{eqn:avg} treats all the relations equally, whereas, as will be shown in the experiments, some relation type is more beneficial for a certain downstream task than others. For example, the co-authorship information between two papers plays a more significant role in predicting the topic of a paper compared with their citation information; eventually, these two information mutually help each other to more accurately predict the topic of a paper.
Therefore, we can adopt the attention mechanism~\cite{bahdanau2014neural} to distinguish between different relation types as follows:
\begin{equation}
\small
\textbf{h}_i=\mathcal{Q}\left(\{\textbf{h}^{(r)}\;|\;r\in\mathcal{R}\}\right)=\sum_{r\in\mathcal{R}}a_i^{(r)}\textbf{h}^{(r)}
\label{eqn:attn}
\end{equation}
where $a_i^{(r)}$ denotes the importance of relation $r$ in generating the final embedding of node $v_i$ defined as:
\begin{equation}
\small
a_{i}^{(r)}=\frac{\exp \left({\textbf{q}}^{(r)} \cdot \mathbf{\textbf{h}}_{i}^{(r)}\right)}{\sum_{r^{\prime}\in\mathcal{R}} \exp \left({\textbf{q}}^{(r^{\prime})} \cdot {\textbf{h}}_{i}^{r^{\prime}}\right)}
\end{equation}
where $\textbf{q}^{(r)}\in\mathbb{R}^d$ is the feature vector of relation $r$. 

\subsubsection{Extension to Semi-Supervised Learning.}
It is important to note that~\proposed~is trained in a \textit{fully unsupervised} manner. However,  in reality, nodes are sometimes associated with label information, which can guide the training of node embeddings even with a small amount~\cite{kipf2016semi,qu2017attention}. To this end, we introduce  a \textit{semi-supervised module} into our framework that predicts the labels of labeled nodes from the consensus embedding \textbf{Z}. More precisely, we minimize the cross-entropy error over the labeled nodes:
\begin{equation}
\ell_{\text{sup}}=-\frac{1}{|\mathcal{Y}_{L}|}\sum_{l \in \mathcal{Y}_{L}} \sum_{i=1}^{c} Y_{l i} \ln \hat{Y}_{l i}
\end{equation}
where $\mathcal{Y}_{L}$ is the set of node indices with labels, $Y\in\mathbb{R}^{n\times c}$ is the ground truth label,
${\hat{Y}}=\textsf{softmax}(f(\textbf{Z}))$ is the output of a softmax layer, and $f:\mathbb{R}^{n\times d}\rightarrow\mathbb{R}^{n\times c}$ is a classifier that predicts the label of a node from its embedding, which is a single fully connected layer in this work. The final objective function with the semi-supervised module is:

\begin{equation}
\mathcal{J}_\textsf{semi}=\sum_{r\in\mathcal{R}}\mathcal{L}^{(r)} + \alpha\ell_{\text{cs}}+ \beta||\Theta|| + \gamma\ell_{\text{sup}}
\label{eqn:semi}
\end{equation}
where $\gamma$ the coefficient of the semi-supervised module.
\section{Experiments}

\noindent\textbf{Dataset. } To make fair comparisons with HAN
~\cite{wang2019heterogeneous}, which is the most relevant baseline method, we evaluate our proposed method on the datasets used in their original paper~\cite{wang2019heterogeneous}, i.e., ACM, DBLP, and IMDB. We used publicly available ACM dataset~\cite{wang2019heterogeneous},
and preprocessed DBLP and IMDB datasets. 
For ACM and DBLP datasets, the task is to classify the papers into three classes (Database, Wireless Communication, Data Mining), and four classes (DM, AI, CV, NLP)\footnote{\textbf{DM}: KDD,WSDM,ICDM, \textbf{AI}: ICML,AAAI,IJCAI, \textbf{CV}: CVPR, \textbf{NLP}: ACL,NAACL,EMNLP}, respectively, according to the research topic. For IMDB dataset, the task is to classify the movies into three classes (Action, Comedy, Drama). 
We note that the above datasets used by previous work are not truly multiplex in nature because the multiplexity between nodes is inferred via intermediate nodes (e.g., ACM: Paper-Paper relationships are inferred via Authors and Subjects that connect two Papers. i.e., ``PAP'' and ``PSP''). Thus, to make our evaluation more practical, we used Amazon dataset~\cite{he2016ups} that genuinely contains a multiplex network of items, i.e., also-viewed, also-bought, and bought-together relations between items.
We used datasets from four categories\footnote{We chose these categories because the three types of item-item relations from these categories are similar in number 
}, i.e., Beauty,  Automotive, Patio Lawn and Garden, and Baby, and
the task is to classify items into the four classes.
For ACM and IMDB datasets, we used the same number of labeled data as in~\cite{wang2019heterogeneous} for fair comparisons, and for the remaining datasets, we used 20 labeled data for each class.
Table~\ref{tab:stats} summarizes the data statistics.

\begin{table*}[t]
	\parbox{.35\linewidth}{
		\small
		\centering
		\caption{Properties of the compared methods (\textit{Mult}.: Mutliplexity, \textit{Attr}: Attribute, \textit{Unsup}: Unsupervised, \textit{Glo}: Global).}
		\renewcommand{\arraystretch}{0.5}
		\label{tab:property}
		\begin{tabular}{p{1.2cm}|cccc}
			& \textit{Mult}. & \textit{Attr}. & \textit{Unsup}. & \textit{Glo}. \\
			\hline
			Dw/n2v & \color{red}{\xmark}     & \color{red}{\xmark}     & \color{blue}{\cmark}     & \color{red}{\xmark} \\
			GCN/GAT & \color{red}{\xmark}     & \color{blue}{\cmark}     & \color{red}{\xmark}     & \color{red}{\xmark} \\
			DGI   & \color{red}{\xmark}     & \color{blue}{\cmark}     & \color{blue}{\cmark}     & \color{blue}{\cmark} \\
			ANRL  & \color{red}{\xmark}     & \color{blue}{\cmark}     & \color{blue}{\cmark}     & \color{blue}{\cmark} \\
			CAN   & \color{red}{\xmark}     & \color{blue}{\cmark}     & \color{blue}{\cmark}     & \color{red}{\xmark} \\
			DGCN  & \color{red}{\xmark}     & \color{blue}{\cmark}     & \color{red}{\xmark}     & \color{blue}{\cmark} \\
			\hline
			CMNA  & \color{blue}{\cmark}     & \color{red}{\xmark}     & \color{blue}{\cmark}     & \color{blue}{\cmark} \\		
			MNE   & \color{blue}{\cmark}     & \color{red}{\xmark}     & \color{blue}{\cmark}     & \color{red}{\xmark} \\
			mGCN & \color{blue}{\cmark}     &  \color{blue}{\cmark}    & \color{blue}{\cmark}     & \color{red}{\xmark} \\
			HAN   & \color{blue}{\cmark}     & \color{blue}{\cmark}     & \color{red}{\xmark}     & \color{red}{\xmark} \\
			\hline
			\proposed  & \color{blue}{\cmark}     & \color{blue}{\cmark}     & \color{blue}{\cmark}     & \color{blue}{\cmark} \\
		\end{tabular}%
	}
\quad
	\parbox{.65\linewidth}{
		\caption{Performance for node clustering and similarity search on test data.}
		\centering
		\small
		\label{tab:clustering}
		\renewcommand{\arraystretch}{0.9}
		\begin{tabular}{p{1.3cm}|>{\centering\arraybackslash}m{0.6cm}>{\centering\arraybackslash}m{0.8cm}|>{\centering\arraybackslash}m{0.6cm}>{\centering\arraybackslash}m{0.8cm}|>{\centering\arraybackslash}m{0.6cm}>{\centering\arraybackslash}m{0.8cm}|>{\centering\arraybackslash}m{0.6cm}>{\centering\arraybackslash}m{0.8cm}}
			\multirow{3}{*}{Method}                           &  \multicolumn{2}{c|}{ACM}                                            & \multicolumn{2}{c|}{IMDB}                                           & \multicolumn{2}{c|}{DBLP}                                           & \multicolumn{2}{c}{Amazon}                                         \\
			\cmidrule{2-9}
			& NMI                              & Sim@5                            & NMI                              & Sim@5                            & NMI                              & Sim@5                            & NMI                              & Sim@5                            \\ 
			\cmidrule{1-9} 
			Deepwalk                     & 0.310                           & 0.710                           & 0.117                           & 0.490                           & 0.348                            & 0.629                           & 0.083                           & 0.726                           \\
			node2vec                     & 0.309                           & 0.710                           & 0.123                           & 0.487                           & 0.382                           & 0.629                           & 0.074                           & 0.738                           \\
			GCN/GAT                      & 0.671                           & 0.867                           & 0.176                           & 0.565                           & 0.465                           & 0.724                           & 0.287                           & 0.624                           \\
			DGI                          & 0.640                           & 0.889                           & 0.182                           & 0.578                           & 0.551                           & 0.786                           & 0.007                           & 0.558                           \\
			ANRL                         & 0.515                           & 0.814                           & 0.163                           & 0.527                           & 0.332                           & 0.720                           & 0.166                           & 0.763                           \\
			CAN                          & 0.504                           & 0.836                           & 0.074                           & 0.544                           & 0.323                           & 0.792                           & 0.001                           & 0.537                           \\
			DGCN                         & 0.691                           & 0.690                           & 0.143                           & 0.179                           & 0.462                           & 0.491                           & 0.143                           & 0.194                           \\ 
			\cmidrule{1-9} 
			CMNA                         & 0.498                           & 0.363                           & 0.152                           & 0.069                           & 0.420                           & 0.511                           & 0.070                           & 0.435                           \\
			MNE                          & 0.545                           & 0.791                           & 0.013                           & 0.482                           & 0.136                           & 0.711                           & 0.001                           & 0.395                           \\
			mGCN                        & 0.668                           & 0.873                           & 0.183                           & 0.550                           & 0.468                           & 0.726                           & 0.301                           & 0.630                           \\
			HAN                          & 0.658                           & 0.872                           & 0.164                           & 0.561                           & 0.472                           & 0.779                           & 0.029                          & 0.495                           \\ 
			\cmidrule{1-9} 
			{\proposed}     & 0.687                           & 0.898                           & \textbf{0.196} & \textbf{0.605} & 0.409                           & 0.766                           & \textbf{0.425} & 0.816                           \\
			{\proposedattn} & \textbf{0.702} & \textbf{0.901} & 0.185                           & 0.586                           & \textbf{0.554} & \textbf{0.798} & 0.412                           & \textbf{0.825} \\
		\end{tabular}
	}
\end{table*}

\smallskip
\noindent\textbf{Methods Compared. }
\vspace{-0.5ex}
\begin{enumerate}[leftmargin=.1in]
	\item[1)] Embedding methods for a single network
	\vspace{-0.5ex}
	\begin{itemize}[leftmargin=.00001in]
			\item \underline{No attributes}: \textbf{Deepwalk}~\cite{perozzi2014deepwalk}, \textbf{node2vec}~\cite{grover2016node2vec}: They learn node embeddings by random walks and skip-gram.
			\item \underline{Attributed network embedding}:	\textbf{GCN}~\cite{kipf2016semi}, \textbf{GAT}~\cite{velivckovic2017graph}: They learn node embeddings based on local neighborhood structures. As they perform similarly, we report the best performing method among them;
					\textbf{DGI}~\cite{velivckovic2018deep}: It maximizes the MI between the graph-level summary representation and the local patches;
					\textbf{ANRL}~\cite{zhang2018anrl}: It uses neighbor enhancement autoencoder to model the node attribute information, and skip-gram model to capture the network structure;
					\textbf{CAN}~\cite{meng2019co}: It learns embeddings of both attributes and nodes in the same semantic space;
					\textbf{DGCN}~\cite{zhuang2018dual}: It models the local and global properties of a graph by employing dual GCNs.
	\end{itemize}
	\item[2)] Multiplex embedding methods
	\vspace{-0.5ex}
	\begin{itemize}[leftmargin=.01in]
		\item \underline{No attributes}: \textbf{CMNA}~\cite{chu2019cross}: It leverages the cross-network information to refine inter-vector for network alignment and intra-vector for other downstream tasks. We use the intra-vector for our evaluations;
			\textbf{MNE}~\cite{zhang2018scalable}: It jointly models multiple networks by introducing a common embedding, and a additional embedding for each relation type.
		\item \underline{Attributed multiplex network embedding}:
			\textbf{mGCN}~\cite{ma2019multi}, \textbf{HAN}~\cite{wang2019heterogeneous}: They apply GCNs, and GATs on multiplex network considering the inter-, and intra-network interactions.
			 For fair comparisons, we initialized the initial node embeddings of mGCN by using the node attribute matrix, although the node attributes information is ignored in the original mGCN;
			\textbf{\proposedattn}: \proposed~with the attention mechanism (Eqn.~\ref{eqn:attn}).
	\end{itemize}
\end{enumerate}

For the sake of fair comparisons with~\proposed, which considers the node attributes, we concatenated the raw attribute matrix $\textbf{X}$ to the learned node embeddings $\textbf{Z}$ of the methods that ignore the node attributes. i.e., Deepwalk, node2vec, CMNA, and MNE. i.e., $\textbf{Z}\leftarrow[\textbf{Z};\textbf{X}]$. Moreover, regarding the embedding methods for a single network, i.e., the methods that belong to the first category in the above list, we obtain the final node embedding matrix $\textbf{Z}$ by computing the average of the node embeddings obtained from each single graph. i.e., $\textbf{Z}=\frac{1}{|\mathcal{R}|}\sum_{r\in\mathcal{R}}\textbf{H}^{(r)}$. 
We provide a summary of the properties of the compared methods in Table~\ref{tab:property}.

\medskip
\noindent\textbf{Evaluation Metrics. }
Recall that~\proposed~is an unsupervised method that does not require any labeled data for training. Therefore, we evaluate the performance of~\proposed~in terms of \textbf{node clustering} and \textbf{similarity search}, both of which are classical performance measures for unsupervised methods. For node clustering, we use the most commonly used metric~\cite{wang2019heterogeneous}, i.e., Normalized Mutual Information (NMI). For similarity search, we compute the cosine similarity scores of the node embeddings between all pairs of nodes, and for each node, we rank the nodes according to the similarity score. Then, we calculate the ratio of the nodes that belong to the same class within top-5 ranked nodes (Sim@5).
Moreover, we also evaluate~\proposed~on the performance in terms of \textbf{node classification}. More precisely, after learning the node embeddings, we train a logistic regression classifier on the learned embeddings in the training set, and then evaluate on the nodes in the test set. 
We use Macro-F1 (MaF1) and Micro-F1 (MiF1)~\cite{wang2019heterogeneous}.

\medskip
\noindent\textbf{Experimental Settings. }
We randomly split our dataset into train/validation/test, and we have the equal number of labeled data for training and validation datasets. We report the test performance when the performance on validation data gives the best result. For~\proposed, we set the node embedding dimension $d=64$, self-connection weight $w=3$, tune $\alpha,\beta,\gamma \in \{0.0001,0.001,0.01,0.1\}$. We implement~\proposed~in PyTorch\footnote{https://github.com/pcy1302/DMGI}, and for all other methods, we used the source codes published by the authors, and tried to tune them to their best performance. More precisely, apart from the guidelines provided by the original papers, we tuned learning rate, and the coefficients for regularization from \{0.0001,0.0005,0.001,0.005\} on the validation dataset. After learning the node embeddings, for fair comparisons, we conducted the evaluations within the same platform.

\begin{table*}[htbp]
	\begin{minipage}{0.65\linewidth}
		\small
		\caption{Node classification performance on test data.}
		\renewcommand{\arraystretch}{0.4}
		\begin{tabular}{p{1.3cm}|cc|cc|cc|cc}
 
			& \multicolumn{2}{c|}{ACM} & \multicolumn{2}{c|}{IMDB} & \multicolumn{2}{c|}{DBLP} & \multicolumn{2}{c}{Amazon} \\
	
			\cmidrule{2-9}      & MaF1 & MiF1 & MaF1 & MiF1 & MaF1 & MiF1 & MaF1 & MiF1 \\
			\cmidrule{1-9}
			Deepwalk & 0.739 & 0.748 & 0.532 & 0.550 & 0.533 & 0.537 & 0.663 & 0.671 \\
			node2vec & 0.741 & 0.749 & 0.533 & 0.550 & 0.543 & 0.547 & 0.662 & 0.669 \\
			GCN/GAT   & 0.869 & 0.870 & 0.603 & 0.611 & 0.734 & 0.717 & 0.646 & 0.649 \\
			DGI   & 0.881 & 0.881 & 0.598 & 0.606 & 0.723 & 0.720 & 0.403 & 0.418 \\
			ANRL   & 0.819	& 0.820 & 0.573&	0.576 &  0.770 & 0.699  & 0.692 &	0.690 \\				
			CAN   & 0.590 & 0.636 & 0.577 & 0.588 &   0.702    &   0.694    & 0.498 & 0.499 \\
			DGCN  & 0.888 & 0.888 & 0.582 & 0.592 & 0.707 & 0.698 & 0.478 & 0.509 \\
			\midrule
			CMNA  & 0.782 & 0.788 & 0.549 & 0.566 & 0.566 & 0.561 & 0.657 & 0.665 \\
			MNE   & 0.792 & 0.797 & 0.552 & 0.574 & 0.566 & 0.562 & 0.556 & 0.567 \\
			mGCN & 0.858 & 0.860 & 0.623 & 0.630 & 0.725 & 0.713 & 0.660 & 0.661 \\
			HAN   & 0.878 & 0.879 & 0.599 & 0.607 & 0.716 & 0.708 & 0.501 & 0.509 \\
			\midrule
			{\proposed}     & \textbf{0.898} & \textbf{0.898} & \textbf{0.648} & \textbf{0.648} & 0.771 & 0.766 & 0.746 & 0.748 \\
			{\proposedattn} & 0.887 & 0.887 & 0.602 & 0.606 & \textbf{0.778} & \textbf{0.770} & \textbf{0.758} & \textbf{0.758} \\
		\end{tabular}%
		\label{tab:classification}%
	\end{minipage}\hfill
	\begin{minipage}{0.3\linewidth}{
			\centering
			\includegraphics[width=\linewidth]{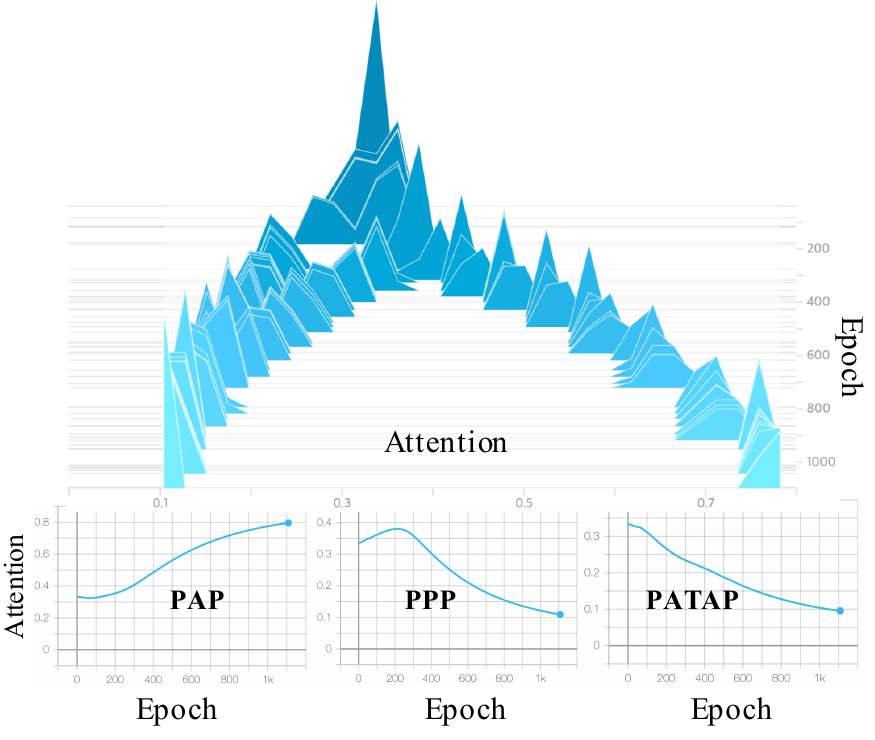}
			\captionof{figure}{Visualization of the attention weights on DBLP dataset.}
			\label{fig:distribution}	
		}
	\end{minipage}
\end{table*}%
\subsection{Performance Analysis}
\noindent\textbf{Overall evaluation. }
Table~\ref{tab:clustering} and Table~\ref{tab:classification} show the evaluation results on unsupervised and supervised task, respectively. We have the following observations: 1) Our proposed~\proposed~and~\proposedattn~outperform all the state-of-the-art baselines not only on the unsupervised tasks, but also the supervised task, although the improvement is more significant in the unsupervised task as expected. This verifies the benefit of our framework that models the multiplexity and the global property of a network together with the node attributes within a single framework.
2) Although DGI shows relatively good performance, the performance is unstable (poor performance on Amazon dataset), indicating that multiple relation types should be jointly modeled.
3) Attribute-aware multiplex network embedding methods, such as mGCN and HAN, generally perform better than those that neglect the node attributes. i.e., CMNA and MNE, even though we concatenated node attributes to the node embeddings. This verifies not only the benefit of modeling the node attributes, but also that the attributes should be systematically incorporated into the model. 4) Multiplex network embedding methods generally outperform single network embedding methods, although the gap is not significant. This verifies that the multiplexity of a network should be carefully modeled, otherwise a simple aggregation of multiple relation-type specific embeddings learned from independent single network embedding methods may perform better.
\begin{table}[t]
	\centering
	\small
	\caption{Performance of similarity search (Sim@5) of embedding methods for a single network. (\textsf{Merged} denotes the average of all the relation-type specific embeddings.)}
	\renewcommand{\arraystretch}{0.1}
	\begin{tabular}{c|c|ccc|cc}
		\toprule
		\multicolumn{2}{c|}{ACM}   &  \begin{tabular}[x]{@{}c@{}}GCN\end{tabular} & DGI   & ANRL  & \multirow{3}[2]{*}{\proposed} & \multirow{3}[2]{*}{\begin{tabular}[x]{@{}c@{}}\proposed \\$_\textsf{attn}$\end{tabular}} \\
		\cmidrule{1-5}
		\multirow{0}[2]{*}{\begin{tabular}[x]{@{}c@{}}Rel.\\Type\end{tabular}} & PAP   & 0.822 & 0.875 & 0.795 &       &  \\
		& PSP   & 0.721 & 0.675 & 0.694 &       &  \\
		\midrule
		\multicolumn{2}{c|}{\textsf{Merged}} & 0.867 & 0.889 & 0.814 & 0.898 & \textbf{0.901} \\
		\midrule
		\midrule
		
		\multicolumn{2}{c|}{IMDB}   &  \begin{tabular}[x]{@{}c@{}}GCN\end{tabular} & DGI   & ANRL  & \multirow{3}[2]{*}{\proposed} & \multirow{3}[2]{*}{\begin{tabular}[x]{@{}c@{}}\proposed \\$_\textsf{attn}$\end{tabular}} \\
		\cmidrule{1-5}
		\multirow{0}[2]{*}{\begin{tabular}[x]{@{}c@{}}Rel.\\Type\end{tabular}} & MAM   & 0.485 & 0.484 & 0.495 &       &  \\
		& MDM   & 0.548 & 0.562 & 0.520 &       &  \\
		\midrule
		\multicolumn{2}{c|}{\textsf{Merged}}& 0.566 & 0.578 & 0.527 & \textbf{0.605} & 0.586 \\
		\midrule
		\midrule
		
		\multicolumn{2}{c|}{DBLP}   & \begin{tabular}[x]{@{}c@{}}GCN\end{tabular} & DGI   & ANRL  & \multirow{4}[2]{*}{\proposed} & \multirow{5}[2]{*}{\begin{tabular}[x]{@{}c@{}}\proposed \\$_\textsf{attn}$\end{tabular}} \\
		\cmidrule{1-5}
		\multirow{1}[3]{*}{\begin{tabular}[x]{@{}c@{}}Rel.\\Type\end{tabular}}& PAP   & 0.730 & 0.779 & 0.692 &       &  \\
		& PPP   & 0.456 & 0.477 & 0.680 &       &  \\
		& PATAP & 0.431 & 0.409 & OOM   &       &  \\
		\midrule
		\multicolumn{2}{c|}{\textsf{Merged}} & 0.724 & 0.786 & 0.720 & 0.766 & \textbf{0.799} \\
		\midrule
		\midrule
		
		\multicolumn{2}{c|}{Amazon} & \begin{tabular}[x]{@{}c@{}}GCN\end{tabular} & DGI   & ANRL  & \multirow{4}[2]{*}{\proposed} & \multirow{5}[2]{*}{\begin{tabular}[x]{@{}c@{}}\proposed \\$_\textsf{attn}$\end{tabular}} \\
		\cmidrule{1-5}
		\multirow{1}[3]{*}{\begin{tabular}[x]{@{}c@{}}Rel.\\Type\end{tabular}}& Also-V & 0.355 & 0.367 & 0.563 &       &  \\
		&  Also-B & 0.357 & 0.381 & 0.516 &       &  \\
		& Bou.-T & 0.662 & 0.639 & 0.770 &       &  \\
		
		\midrule
		\multicolumn{2}{c|}{\textsf{Merged}} & 0.624 & 0.558 & 0.764 & 0.816 & \textbf{0.825} \\
		\bottomrule
	\end{tabular}%
	\label{tab:attn}%
\end{table}%

\medskip
\noindent\textbf{Effect of the attention mechanism. }
In Table~\ref{tab:attn}, we show the performance of~\proposed~and~\proposedattn, together with the performance of  single network embedding methods (GCN/GAT, DGI, and ANRL). We observe that~\proposedattn~outperforms~\proposed~in most of the datasets but IMDB dataset. To analyze the reason for this, we first plot the distribution of the attention weights on DBLP dataset over the training epochs in Figure~\ref{fig:distribution}. The above graph in Figure~\ref{fig:distribution} demonstrates that the attention weights eventually end up in both extremes. i.e., close to 0 or close to 1, and the below graphs show that most of the attention weight is dedicated to a single relation type, i.e., ``PAP'', which actually turns out to be the most important relation among the three (See Table~\ref{tab:attn}); This phenomenon is common in every dataset.
Next, we look at the performance of the single network embedding methods, especially DGI, on each relation type in Table~\ref{tab:attn}. We observe that the performance differences among relation types in ACM, DBLP, and Amazon datasets are more biased to a single relation type, whereas in IMDB dataset, ``MAM'' and ``MDM'' relations relatively show similar performance.
To summarize our findings, since the attention mechanism tends to favor the single most important relation type (``PAP'' in ACM, ``MDM'' in IMDB, ``PAP'' in DBLP, and ``Bought-together'' in Amazon),~\proposedattn~outperforms~\proposed~on datasets where one relation type significantly outperforms the other, i.e., ACM, DBLP, and Amazon, by removing the noise from other relations. On the other hand, for datasets where all the relations show relatively even performance, i.e., IMDB, extremely favoring a single well performing relation type (``MDM'') is rather detrimental to the overall performance because the relation ``MAM'' should also be considered to some extent. 

We also note that since the attention mechanism of~\proposedattn~can infer the importance of each  relation type, we can filter out unnecessary relation types as a preprocessing step. To verify this, we evaluated on all possible combinations of relation types in DBLP dataset (Table~\ref{tab:attnfilter}). We observe that by removing the relation ``PATAP'', which turned out to be the most useless relation type in Table~\ref{tab:attn},~\proposedattn~obtains even better results than using all the relation types, whereas for GCN and DGI, still considering all the relation types shows the best performance. This indicates that the attention mechanism can be useful to filter out unnecessary relation types, which will especially come in handy when the number of relation types is large.

\begin{table}[t]
	\centering
	\small
	\renewcommand{\arraystretch}{0.4}
	\caption{NMI on various combinations of relation types.}
	\begin{tabular}{c|l|ccc}
		\multicolumn{2}{c|}{DBLP dataset} & \multicolumn{1}{l}{GCN/GAT} & \multicolumn{1}{l}{DGI} & \multicolumn{1}{l}{\proposedattn} \\
		\midrule
		\multirow{5}[2]{*}{NMI} & PAP+PPP & 0.464 & 0.543 & \textbf{0.565} \\
		& PAP+PATAP & 0.458 & 0.535 & 0.017 \\
		& PPP+PATAP & 0.332 & 0.237 & 0.201 \\
		\cmidrule{2-5}
		& \textit{All}   & \textbf{0.465} & \textbf{0.551} & 0.554 \\
	\end{tabular}%
	\label{tab:attnfilter}%
\end{table}%

\smallskip
\noindent\textbf{Ablation study. }
To measure the impact of each component of~\proposedattn, we conduct ablation studies on the largest dataset, i.e., DBLP,  in Table~\ref{tab:ablation}. We have the following observations: 1) As expected, the semi-supervised module specifically helps improve the node classification performance, which is a supervised task, whereas the performance on the unsupervised task remains on par.  2) Various readout functions including ones that contain trainable weights (Linear projection and SAGPool~\cite{lee2019self}) do not have much impact on the performance, which promotes our use of average pooling. 3) The second term in Eqn.~\ref{eqn:reg} indeed plays a significant role in the consensus regularization framework. 4) The sharing of the scoring matrix $\textbf{M}$ facilitates~\proposed~to model the interaction among multiple relation types. 5) Node attributes are crucial for representation learning of nodes. 6) Shuffling adjacency matrix instead of attribute matrix deteriorates the model performance. 

\begin{table}[t]
	\centering
	\small
	\caption{Result for ablation studies of~\proposedattn.}
	\renewcommand{\arraystretch}{0.75}
	\begin{tabular}{cc|c|c|c}
		\multicolumn{2}{c|}{DBLP dataset} & MaF1 & NMI   & Sim@5 \\
		\hline
		\multicolumn{2}{c|}{\proposedattn} & 0.778 & 0.554 & 0.798 \\
		\hline
		\multicolumn{2}{l|}{1) \proposedattn + Semi supervised} & 0.791 & 0.555 & 0.798 \\
\hline
		\multicolumn{1}{l|}{\multirow{5}[0]{*}{2) \begin{tabular}[x]{@{}c@{}}\textsf{Readout}\\(Eqn.~\ref{eqn:readout})\end{tabular}}} & Random sample & 0.774 & 0.555 & 0.797\\
		\multicolumn{1}{l|}{} & Maxpool & 0.778 & 0.552 & 0.802 \\
		\multicolumn{1}{l|}{} & Linear projection & 0.783 & 0.565 & 0.803 \\
		\multicolumn{1}{l|}{} & SAGPool & 0.797 & 0.563 & 0.797 \\
		
		\hline
		\multicolumn{2}{l|}{3) Without 2nd term of Eqn.~\ref{eqn:reg}} & 0.749 & 0.448 & 0.787 \\
		\multicolumn{2}{l|}{4) $\textbf{M}\neq\textbf{M}^{(1)}\neq...\neq\textbf{M}^{(|\mathcal{R}|)}$.} & 0.645 & 0.076 & 0.677  \\
		\multicolumn{2}{l|}{5) No attributes (Adj. as attribute)} & 0.377 & 0.053 & 0.763 \\
		\multicolumn{2}{l|}{6) Neg sample: Shuffle adj.} & 0.364 & 0.156 & 0.504 \\
	\end{tabular}%
	\label{tab:ablation}%
\end{table}%

\section{Related Work}
\noindent\textbf{Network embedding. } 
Network embedding methods aim at learning low-dimensional vector representation for nodes in a graph while preserving the network structure~\cite{perozzi2014deepwalk,grover2016node2vec,tang2015line}, and various other properties such as node attributes~\cite{zhang2018anrl,meng2019co}, structural role~\cite{ribeiro2017struc2vec},
and node label information~\cite{huang2017label}. 

\smallskip
\noindent\textbf{Multiplex Network embedding. } 
A multiplex network, which is also known as a multi-view network~\cite{tang2015line,shi2018mvn2vec} or a multi-dimensional network~\cite{ma2018multi,ma2019multi} in the literature, consists of multiple relation types among a set of single-typed nodes. It can be thought of as a special type of heterogeneous network~\cite{dong2017metapath2vec,fu2017hin2vec} with a single type of node and multiple types of edges. Therefore, a multiplex network calls for a special attention because there is no need to consider the semantics between different types of nodes, which is often addressed by the concept of meta-path~\cite{sun2011pathsim}. Distinguished from heterogeneous network, a key challenge in the multiplex network embedding is to learn a consensus embedding for each node by taking into account the interrelationship among the multiple graphs. In this regard, existing methods mainly focused on how to integrate the information from multiple graphs.
HAN~\cite{wang2019heterogeneous} employed graph attention network~\cite{velivckovic2017graph} on each graph, and then applied the attention mechanism to merge the node representations learned from each graph by considering the importance of each graph. 
However, the existing methods either require labels for training~\cite{wang2019heterogeneous,qu2017attention,schlichtkrull2018modeling}, or overlook the node attributes~\cite{liu2017principled,xu2017multi,li2018multi,shi2018mvn2vec,zhang2018scalable,ni2018co,chu2019cross}.
Most recently, \citet{ma2019multi} proposed a graph convolutional network (GCN) based method called mGCN, which is not only unsupervised, but also naturally incorporates the node attributes by using GCNs. However, since it is based on GCNs that capture the local graph structure~\cite{yadav2019lovasz}, it fails to fully model the global properties of a graph~\cite{zhuang2018dual,wang2016structural,velivckovic2018deep}.

\smallskip
\noindent\textbf{Attributed Network Embedding. } 
Nodes in a network are often affiliated with various contents, such as abstract text in the publication network, user profiles in social networks, and item description text in movie database or item networks. Such networks are called attributed networks, and have been extensively studied~\cite{li2017attributed,hamilton2017inductive,yang2015network,zhang2018anrl,gao2018deep,zhou2018prre,velivckovic2018deep,meng2019co}. Their goal is to preserve not only the network structure, but also the node attribute proximity in learning representations.
Recently, GCNs~\cite{kipf2016semi,velivckovic2017graph,velivckovic2018deep} have been widely praised for its seamless integration of the network structure, and node attributes into a single framework.

\smallskip
\noindent\textbf{Mutual Information. } 
it has been recently made possible to compute the MI
between high dimensional input/output pairs of deep neural networks~\cite{belghazi2018mine}. Several recent work adopted the infomax principle~\cite{linsker1988self} to learn the unsupervised representations in different domains, such as images~\cite{hjelm2018learning}, speech~\cite{ravanelli2018learning} and graphs~\cite{velivckovic2018deep}. More precisely, 
~\citet{velivckovic2018deep} proposed Deep Graph Infomax (DGI) for learning representations of graph structured inputs by maximizing the MI between a high-level global representation, and the local patches of a graph.

\section{Conclusion}
We presented a simple yet effective unsupervised method for embedding attributed  multiplex network.~\proposed~can jointly integrate the embeddings from multiple types of relations between nodes through the consensus regularization framework, and the universal discriminator. Moreover, the attention mechanism of~\proposedattn~can infer the importance of each relation type, which facilitates the preprocessing of the multiplex network. Experimental results on not only unsupervised tasks, but also a supervised task verify the superiority of our proposed framework. 

\fontsize{9.0pt}{10.0pt} \selectfont
\bibliography{Bibliography-File}
\bibliographystyle{aaai}
\end{document}